\documentclass[conference]{IEEEtran}
\IEEEoverridecommandlockouts
\usepackage{cite}
\usepackage{amsmath,amssymb,amsfonts}
\usepackage{algorithmic}
\usepackage{graphicx}
\usepackage{textcomp}
\usepackage{booktabs}
\usepackage{xcolor}
\usepackage{hyperref}
\def\BibTeX{{\rm B\kern-.05em{\sc i\kern-.025em b}\kern-.08em
    T\kern-.1667em\lower.7ex\hbox{E}\kern-.125emX}}
\begin{document}

\title{COVID-19 Face Mask Recognition with Advanced Face Cut Algorithm for Human Safety Measures}

\author{\IEEEauthorblockN{Arkaprabha Basu1}
\IEEEauthorblockA{\textit{Electronics and Communication Sciences Unit} \\
\textit{Indian Statistical Institute}\\
Kolkata, India \\
arkaprabha17@gmail.com}
\and
\IEEEauthorblockN{Md Firoj Ali2}
\IEEEauthorblockA{\textit{Assistant Professor, Department of Computer Science} \\
\textit{Ramakrishna Mission Residential College, Narendrapur}\\
Kolkata, India \\
firojali.mca@gmail.com}}

\maketitle

\begin{abstract}
In the last year, the outbreak of COVID-19 has deployed computer vision and machine learning algorithms in various fields to enhance human life interactions. COVID-19 is a highly contaminated disease that affects mainly the respiratory organs of the human body. We must wear a mask in this situation as the virus can be contaminated through the air and a non-masked person can be affected. Our proposal deploys a computer vision and deep learning framework to recognize face masks from images or videos. We have implemented a Boundary dependent face cut recognition algorithm that can cut the face from the image using 27 landmarks and then the preprocessed image can further be sent to the deep learning ResNet50 model. The experimental result shows a significant advancement of 3.4 percent compared to the YOLOV3 mask recognition architecture in just 10 epochs.
\end{abstract}


\section{Introduction}
In the year 2020, the world faced a new challenge which was due to a virus that was killing people on a large scale. There were 138,057,338 cases and 2,972,992 deaths worldwide till the date due to the COVID-19 virus. COVID-19 is a highly infectious disease that comes with major symptoms of fever, dry cough, tiredness, and breathing problems most of the time. The minor symptoms can be diagnosed as diarrhea, conjunctivitis, loss of taste or smell, rash on the skin, or discoloration of fingers or toes.  The technology and medical sciences have improved continuously to adapt the treatment of this disease but that hasn't stopped the spread of virus\cite{1,2,3}.Very recently India, Brazil, USA is facing the second wave of COVID-19 and as the virus is mutating very fast nowadays it is becoming difficult to detect affected ones from outside as in most cases the people are asymptomatic or very less symptomatic.

The one and the only solution is to take measurements so that the virus can't affect others or can't also enter into the body\cite{4}. The main body parts which can be contamination points are fingers, nose, and face. To stop the contamination from finger we need to use sanitiser most of the time we touch something mostly in public areas. The face mask is very useful while in the public area to stop contamination through the air. The people who are wearing a face mask have less chance to get ill by the virus than the non-masked people. People with or without face masks need to be identified as it is important for future contamination or outbreak in a family or area. Additionally, a red area with a larger number of cases and non-masked people should be alerted for a fine or legal action as the disease can affect them without a face mask.

This paper empowers automatic detection of face masks from images or videos using computer vision and deep learning at the backend. The trained model with updated preprocessing rather than the conventional preprocessing techniques have been tested on Dataset to make a clear decision when seeing a human face. Several papers have been published recently to detect face masks from images.  The YOLOV3,YOLOV5\cite{15} model has been proposed to classify face mask from the images\cite{5}. YOLO is a very important vision learning tool that can be applied to almost every place or mode\cite{6,7}. This efficient learning technique is well documented in the field of face recognition, real-time object recognition\cite{8,14}. Our proposal comes with the idea of CNN into this task\cite{9}. With the CNN model of ResNet50, which is known for its vast architecture, we have worked with face fiducial point detection and cut the face(Figure 1) from the total scenery so that model can get only the face, not the other areas.
\begin{center}
    \includegraphics[width=4cm, height=3cm]{"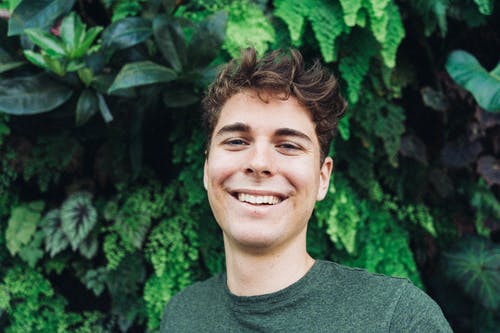"}
    \includegraphics[width=3cm, height=3cm]{"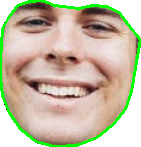"}
    \includegraphics[width=4cm, height=3cm]{"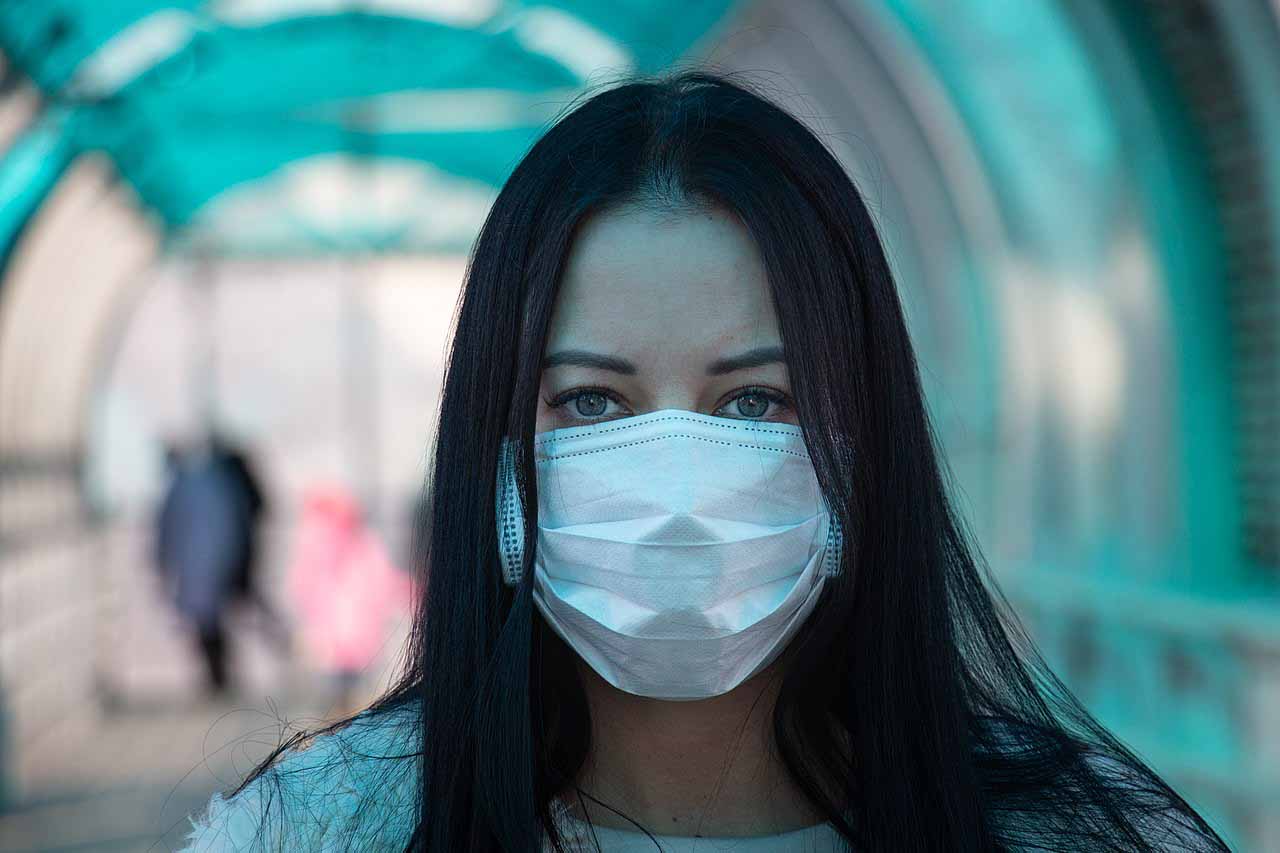"}
    \includegraphics[width=3cm, height=3cm]{"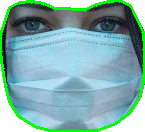"}
    
    \begin{footnotesize}
    Figure-1 Face cut method applied on the pictures
\end{footnotesize}

\end{center}

\section{Method Description}

\subsection{Dataset}
Dataset is one of the most important part to train a deep learning model. In the case of the dataset, we have explored 3 publicly available datasets. Our proposal applies the preprocessing technique to the dataset and the output results are directly sent to the deep learning model thereafter. In the case of the first dataset, we are motivated by the work of Prajna Bhanadary's original dataset of face mask\cite{10}. These are manually edited with a face mask to some popular pictures of celebrities. The second dataset is the fine-tuning dataset which works better than the first one. This dataset has been taken from Kaggle\cite{11} and has almost 7500 files with and without the mask. The third dataset consists of almost 11,792 images of two categories of masked and non-masked\cite{12}. This dataset is a detailed dataset that can add more fresh entries into the pre-processing and increase the efficiency of the model as it is a CNN model, and the training portion needs lots of data. The total dataset adds almost 20,721 fresh entries to the preprocessing technique so that it can cut the face from those irrespective of the picture light-dark ratio or the picture of the face is one-sided or frontal.
\begin{center}
\begin{footnotesize}
Table 1 Dataset Used for the model Construction
    \end{footnotesize}
\\[.1in]
\begin{tabular}{c|c|c|c|c}
\toprule
Dataset & Source & Without Mask & With Mask & Total\\
\midrule
A & Github & 686 & 690 & 1,376 \\
B & Kaggle & 3,828 & 3,725 & 7,553\\
C & Kaggle & 5,909 & 5,883 & 11,792\\
\bottomrule
Total & -- & 10,423 & 10,298 & 20,721\\
\bottomrule
\end{tabular}
\end{center}

\begin{center}
    \includegraphics[width=4cm, height=3cm]{"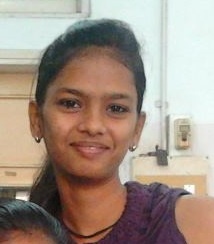"}
    \includegraphics[width=4cm, height=3cm]{"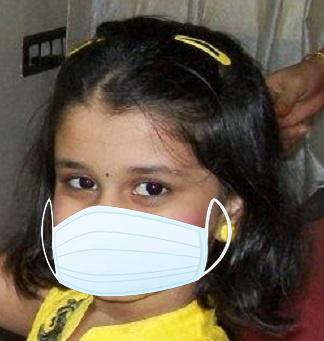"}
    \includegraphics[width=4cm, height=3cm]{"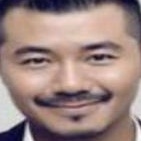"}
    \includegraphics[width=4cm, height=3cm]{"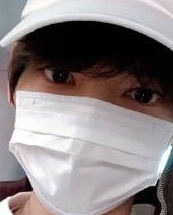"}
    \includegraphics[width=4cm, height=3cm]{"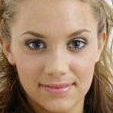"}
    \includegraphics[width=4cm, height=3cm]{"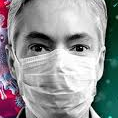"}
    
    \begin{footnotesize}
    Figure-2 Samples from Dataset A, Dataset B and Dataset C
\end{footnotesize}

\end{center}

\subsection{Preliminary Processing}
Before we dive into the model and recognition of face masks from images, the most important part of the proposal is the preliminary processing. The paper describes the preliminary processing part with some steps which have motivated the ultimate preprocessing for the model.

To make the ultimate model we are only concerned about the faces from the picture. Our method uses Python 3.0 as the programming language and DLIB package for this task. This package helps us to recognize the face from the image(Figure 3). The face cut method encourages the task of mask recognition in further steps.

\begin{center}
    \includegraphics[width=4cm, height=3cm]{"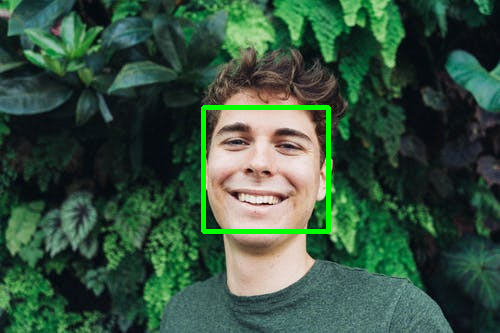"}
    \includegraphics[width=4cm, height=3cm]{"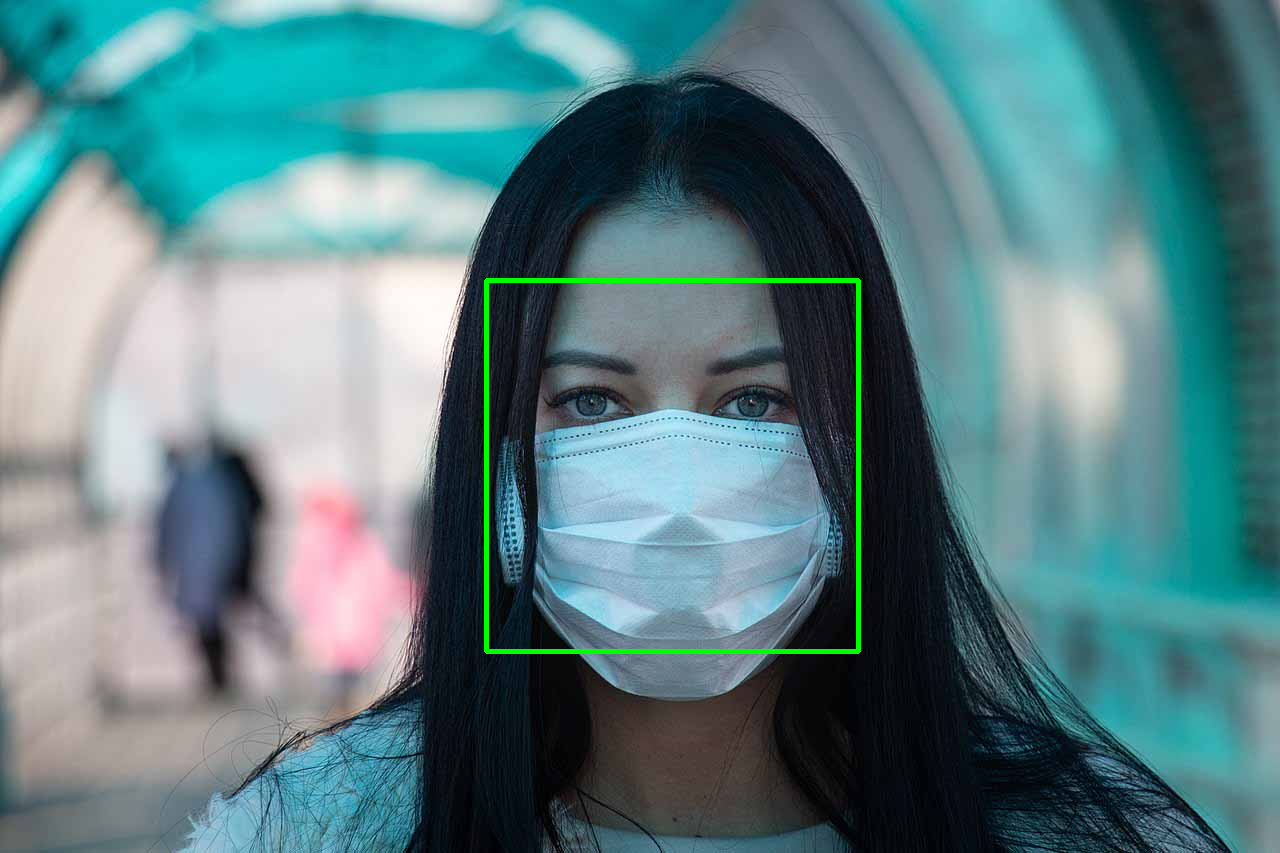"}
    \begin{footnotesize}
    Figure-3 Face recognition from the samples
\end{footnotesize}

\end{center}

The next stage of the proposal comes with the hand of face points detection from the recognized faces. This part is the most important one as the face cut algorithm is dependent on the 67 points recognized by this part. This algorithm tries to find out most of the points from the faces using the 2D distance from the nose point. The model for this task has been trained on various images where the face points have been recognized by the euclidian geometrical distance from the nose point to the upper, left, right, lower boundaries of the face. The pre-trained model we have used from \url{https://github.com/italojs/facial-landmarks-recognition/blob/master/shape_predictor_68_face_landmarks.dat}. This pre-trained model predicts facial landmarks from the face and points to those within the image. For the face masked image as the points can't be recognized so easily, it generates the predicted points which are quite satisfactory. The model detects a series of points from 1-67. The points represent different positions of the faces.
\begin{center}
Jaw points - 0–16\\
Right brow points - 17–21\\
Left brow points - 22–26\\
Nose points - 27–35\\
Right eye points - 36–41\\
Left eye points - 42–47\\
Mouth points - 48–60\\
Lips points - 61–67\\
\end{center}
Normal face recognition uses the bounding box to detect the face point. The pre-trained model recognizes landmark points from the face. Our optimal preprocessing technique deals with the points detected by the model and so that we can fine-tune the points and use them to cut only the bounded positions from the actual image.

Our proposal works with only the boundaries of the face from the face points and removes all the unnecessary parts of the pictures as they can't contribute to the face mask recognition task. The boundary of the face can be recognized using jaw points, eyebrow points, and uppermost nose points. We have used 1-16 face jaw points from the image as those contribute to the boundary of the bottom face portion. The points which have been picked up by the model in 1-16 order that recognizes the face from right ear to left ear jaw points. The only points needed after the 1-16 points are those which contribute to the upperside boundary of the faces. For this task, the method picks up the 17 -26 point and only then the 27th as this will be the connecting point between 17-21 and 22-26. Later we have appended all the points in a list and plotted those into the actual image. These eliminate all the inside points of the eyes, nose, lips as we need only the boundaries. The list of the needed points has been used to draw a circle with the points that create the face area to be cut. Figure 4 describes all steps of preprocessing needed for the ultimate cut.

\begin{center}
    \includegraphics[width=4cm, height=3cm]{"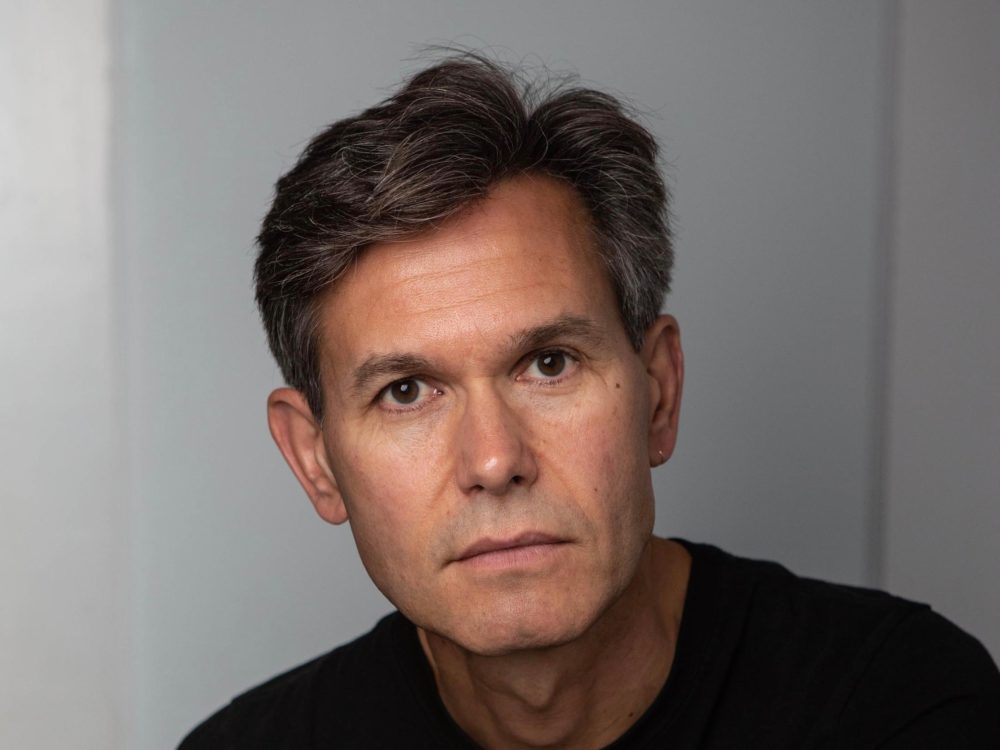"}
    \includegraphics[width=4cm, height=3cm]{"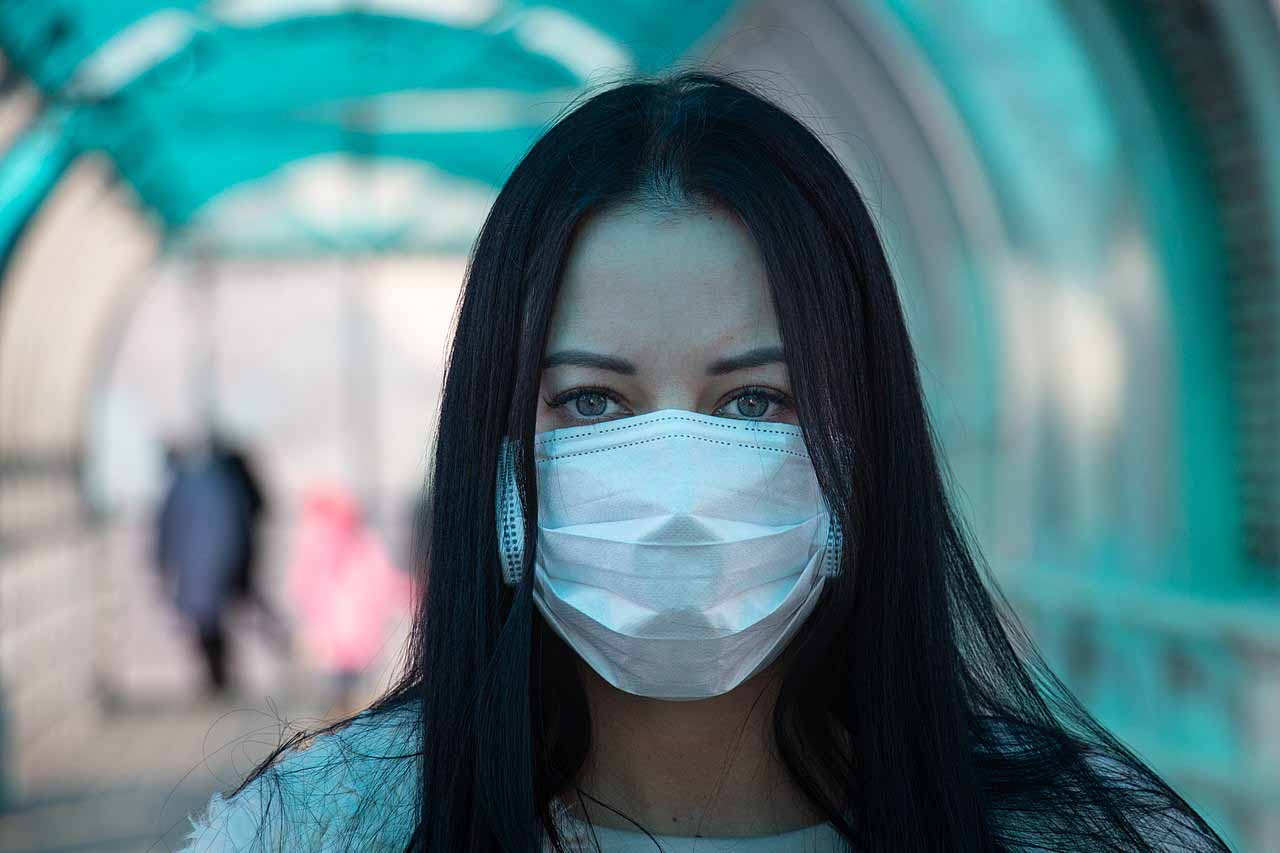"}
    \includegraphics[width=4cm, height=3cm]{"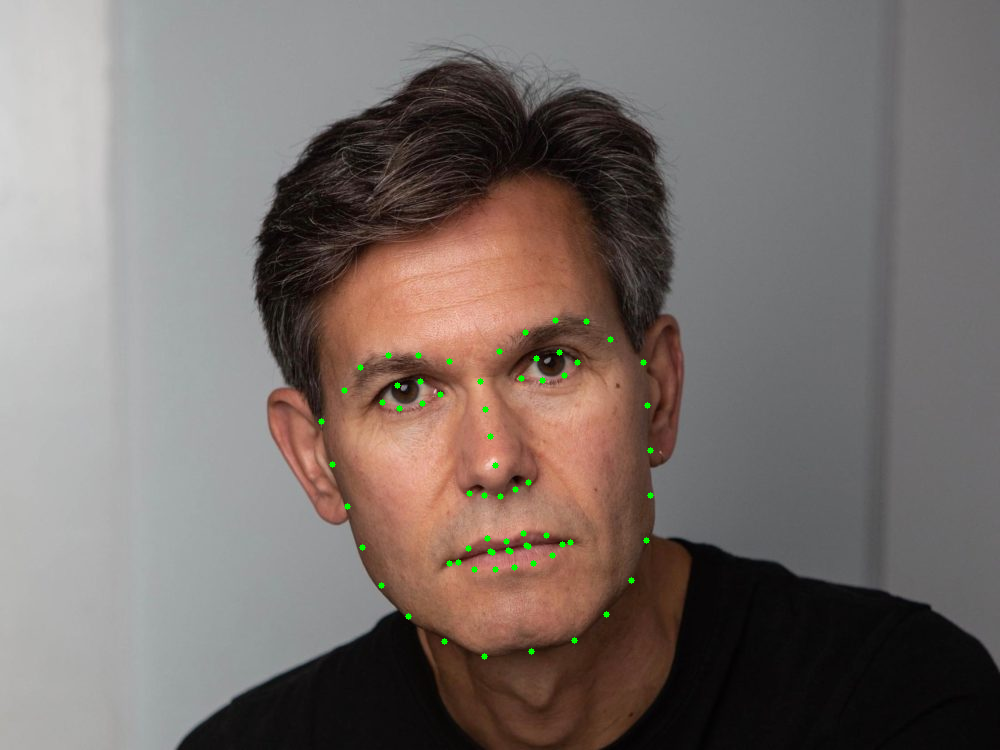"}
    \includegraphics[width=4cm, height=3cm]{"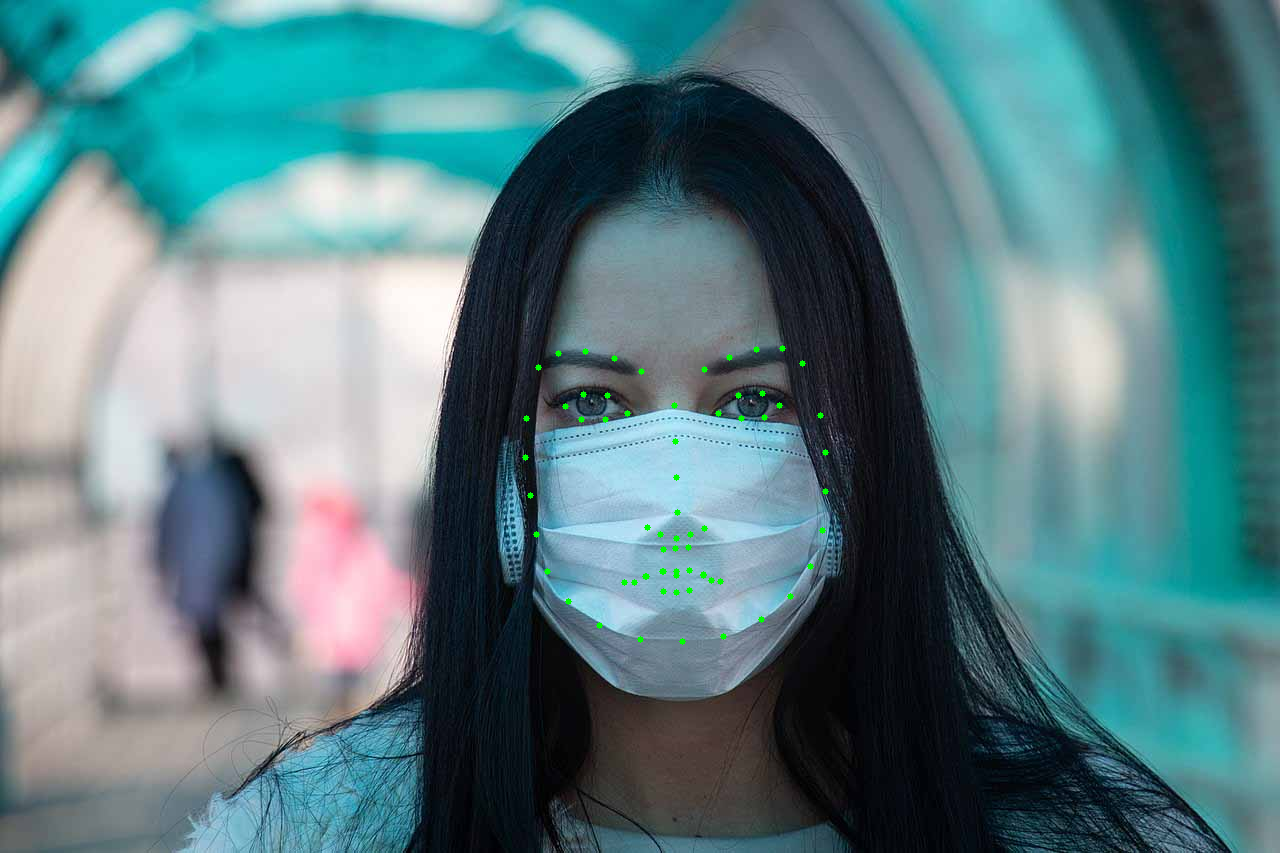"}
    \includegraphics[width=4cm, height=3cm]{"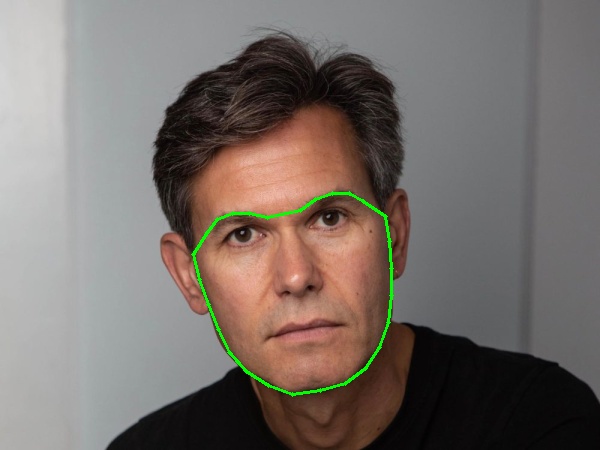"}
    \includegraphics[width=4cm, height=3cm]{"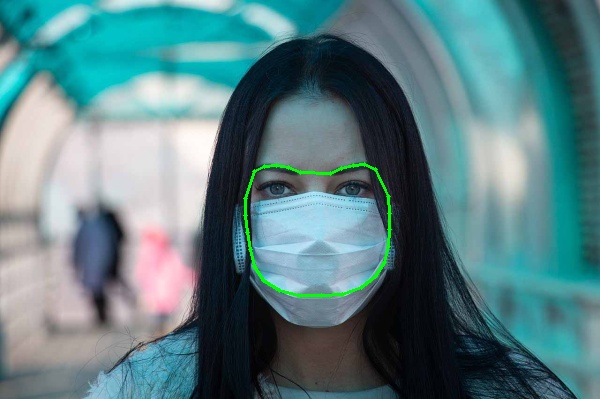"}
    \includegraphics[width=4cm, height=3cm]{"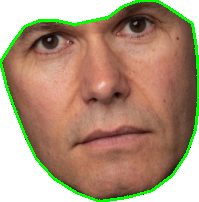"}
    \includegraphics[width=4cm, height=3cm]{"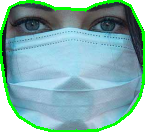"}
    
    \begin{footnotesize}
    Figure-4 Preprocessing steps for Ultimate Face cut algorithm
\end{footnotesize}

\end{center}

The ultimate cut from the image has been carried out through the total dataset that creates the ultimate cut dataset and further can be used for the model prediction. This technique can be termed as a data augmentation technique as we use the term in case of less data availability or to get actual data points. This method is far more powerful than normal augmentation techniques like zooming, shearing, cropping as they are blind to know which part of the image is needed. Our method outperforms those by predicting the face part from the image using a pre-trained model in the pre-processing task.

\subsection{Model Creation }
After the pre-processing for the model prediction, we have used the largely known ResNet50 model. This model is a model of 48 layers, one max-pool, and one average pool layer. This model is the winner of the 2015 ImageNet and MS-COCO competition as this was obtained as the best model for accuracy. The main idea of this model is to use a dense layer at the very last stage to detect the flatten feature vector from the points. This paper used a ResNet50 model and imported the weights of ImageNet to make the model pre-trained by it. The ImageNet is the largest database with 1000 categories of images of animals and objects around us. 

In case of our task, we will have 2 classes for recognition. For this, we have added a global average pool layer followed by a dense layer and rejected the last layer of actual ResNet50. We have taken number of classes for the last layer detection task and predicted only those in the last layer of the model. This work encourages the work of Domain Adaptation\cite{13} in a very short term as domain adaptation adds or subtracts one matured model to get a higher prediction for the new task.
\begin{center}
    \includegraphics[width=8cm, height=6cm]{"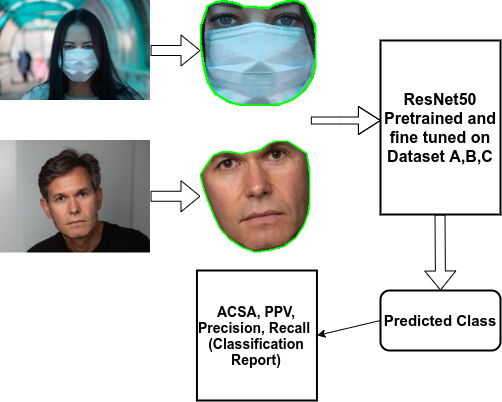"}
    
    \begin{footnotesize}
    Figure-5 Model architecture for Testing
\end{footnotesize}

\end{center}
Every model learns some of the distinct features for each class so that in the testing phase it can use some of those to make the recognition. This paper adds a softmax layer at the last stage so that we can remove it to get the feature embeddings to plot the visualization from the model in the later stage of the result. At the very first stage of pre-processing, we are forcing the model to learn only from the face by cutting it using computer vision techniques of a pre-trained model. The model trains itself with 60 20 20 splits in training validation and testing with 10 epochs and learns the feature vectors only from the face. In the later stage of testing, we give it actual pictures to learn the points from the face and make the recognition.
\section{Results}
The dataset has been processed through the pre-trained ResNet50 model to make the predictions. The resulting stage is the reflection of the prediction which supports the efficiency of the created architecture in various terms of Accuracy and loss. 

In the result stage, the paper discusses the Test Accuracy, Average Class-specific accuracy(ACSA), Predicted Positive Value(PPV), loss value from the model. We have explored the model with Kullback-Leibler loss and cross-entropy loss while the cross-entropy results in better accuracy in the testing stage. We have explored the total classification report after detection and calculated various parameters from it mentioned above. The ACSA has been taken by a mean of two classes in the detection stage. PPV is an important parameter that checks the number of samples of class A that has been detected as class B and vice versa. All values have been calculated from the confusion matrix(COMA) and classification report. Equation 1, 2, 3 describes the mathematical formula of these parameters used.

\begin{center}
\begin{equation}
ACSA = \frac{1}{N}\sum_{i=1}^{N}\frac{COMA[i,i]}{COMA[i,:]}
\end{equation}
\end{center}
\begin{center}
\begin{equation}
Class Accuracy=list[\frac{COMA[i,i]}{COMA[i,:]}]
\end{equation}
\end{center}
\begin{center}
\begin{equation}
PPV=list[\frac{COMA[i,i]}{COMA[:,i]}]
\end{equation}
\end{center}[.2in]
The class accuracy comes from the number of samples passed into the model and has been detected as the right class while the PPV deals with the number of samples passed and the number of samples not detected as other classes. Both are important but PPV deals with the correlation of the classes between the dataset. More a class has been detected as the other one means the similarity between those two classes. 

Our paper compares some of the well-known architectures which have been published in recent years and outperformed those. The YOLOV3\cite{5} architecture has been trained with 4000 epochs to get 96 percent accuracy and 0.0730 loss. YOLOV5\cite{15} is an advancement on the previous architecture and achieved almost 97.9 percent accuracy. There are much more architecture and machine learning models that have introduced Support Vector Machine(SVM), K Nearest Neighbor(KNN), MobileNet for the face mask recognition task as this task is important for the future outbreak of pandemic diseases. 

We have compared our model with the ResNet50, YOLOV3, YOLOV5, SVM, KNN, MobileNet in table 2. Later table 3, 4 describes various parameters of accuracy and loss function which has been calculated from the best model. This comparsion table clearly mentions the superiority of the advanced preprocessing technique proposed in the paper.
\begin{center}
    
\begin{footnotesize}
Table 2 Result Calculated for Various Pre-processing techniques on Data sets
    \end{footnotesize}
\\[.1in]
   \begin{tabular}{c|c|c}
    \toprule
    Method & Accuracy & Loss\\
    \midrule
    Support Vector Machine(SVM)  & 89.4 & - \\
    K-Nearest Neighbor(KNN) & 87.8 & -\\
    MobilNet & 94.2 & -\\
    YOLOV3 & 96 & .0730\\
    YOLOV5 & 97.9 & -\\
    ResNet50 & 97.2 & .0578\\
    Face Cut + ResNet50 & \textbf{99.4} & .0183\\
\bottomrule
    \end{tabular}
    \end{center}

We have calculated the Confusion matrix for the proposed model and described how the model has recognized the binary classification objects. From this matrix, we have calculated the accuracy parameters later in table 3.
\begin{center}
\begin{equation}
    \begin{pmatrix} 
       2058 & 26\\
        11 & 2048
    \end{pmatrix}. 
\end{equation}
\end{center}
Figure 6 mentions the validation and training plotting with the technique. The plotting mentions the validation and training accuracy, loss as it has done the testing stage later on test-generator. We have used OpenCV for preprocessing, and TensorFlow for importing the ResNet50 model. The confusion matrix mentions the efficiency of the model on the testing stage.
\begin{center}
    \includegraphics[width=4cm, height=3cm]{"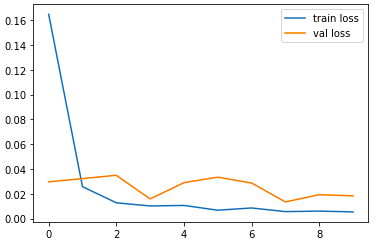"}
    \includegraphics[width=4cm, height=3cm]{"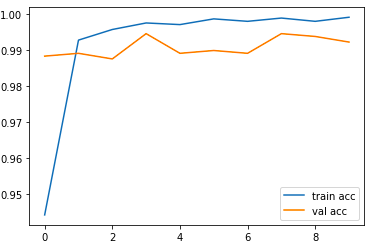"}
    \begin{footnotesize}
    Figure-6 Loss and Accuracy after the model training
\end{footnotesize}
\\[.1in]
\begin{footnotesize}
Table 3 Accuracy Calculation for the Proposed Model
    \end{footnotesize}
\\[.1in]
   \begin{tabular}{c|c|c|c|c}
    \toprule
    Class & Precision & Recall & F1-Score & Support\\
    \midrule
    Without Mask & 1.00 & .99 & .99 & 2084 \\
    With Mask & .99 & 1.0 & .99 & 2059 \\
    \bottomrule
    Accuracy & & &.99 & 4143\\
    Macro Average & .99 & .99 & .99 & 4143\\
    Weighted Average & .99 & .99 & .99 & 4143\\
\bottomrule
    \end{tabular}
    \\[.1in]
\begin{footnotesize}
Table 4 Accuracy Calculation for the Proposed Model
    \end{footnotesize}
\\[.1in]
   \begin{tabular}{c|c|c|c}
    \toprule
    Parameter & Without Mask & With Mask & Average\\
    \midrule
    Class Accuracy &0.9979 &0.9897\\
    ACSA & & & .9938\\
    PPV &0.9883, &0.9959\\
\bottomrule
    \end{tabular}
    \end{center}
\subsection{Visualization}
The last layer we have used for the ResNet50 model has an activation input of Softmax. The softmax makes an accurate prediction of the feature embedding positions and calculates the maximize class value predicted by the model. By removing the softmax activation function we can get the feature embedding on a NumPy array. The array can further be resized to the picture size so that we can recognize a heatmap from the embedding. Those heatmaps can further be processed for the Grad-CAM visualization on the test image. The model has been run on some sample images and Grad-CAM visualization and prediction have been recorded..

\begin{center}
    \includegraphics[width=4cm, height=3cm]{"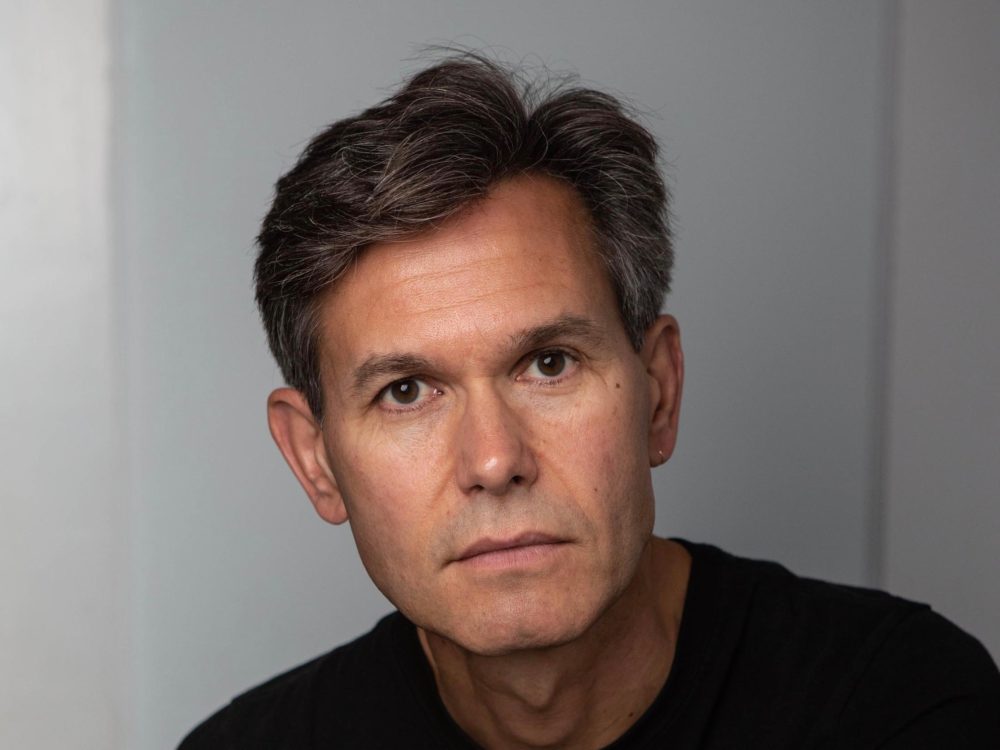"}
    \includegraphics[width=4cm, height=3cm]{"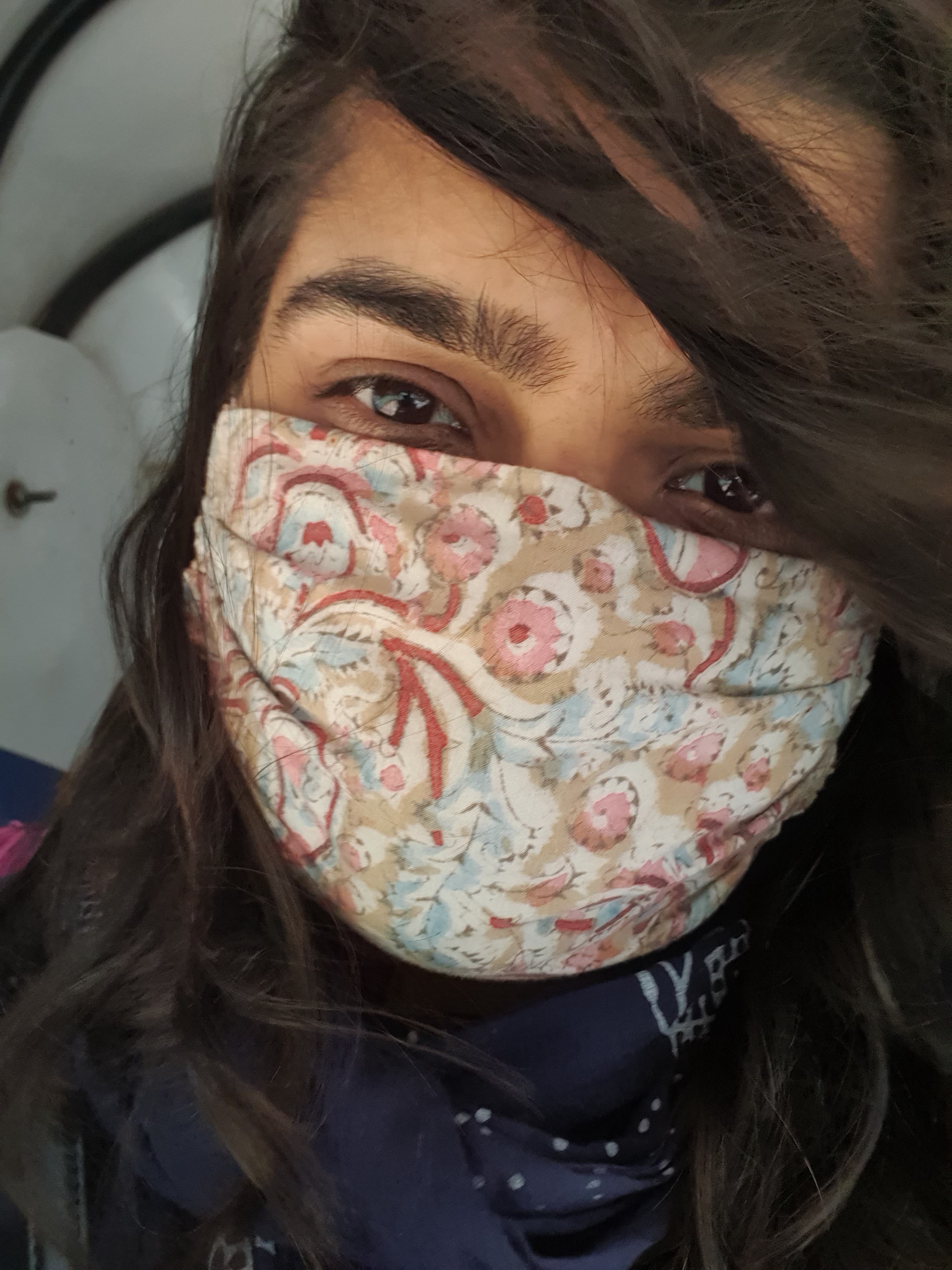"}
    \includegraphics[width=4.2cm, height=3.5cm]{"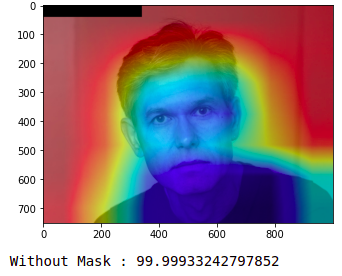"}
    \includegraphics[width=4.2cm, height=3.5cm]{"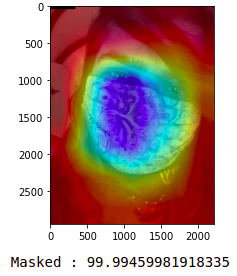"}
    \begin{footnotesize}
    Figure-7 GRAD-CAM visualization on Mask and Without Mask image
\end{footnotesize}
\end{center}
The Explainable AI technique proposed in the method not only shows the accuracy of the masked or non-masked person rather than also shows the Feature points that have been taken for the model to make the decision. The models which have previously explained work with a large number of dense inputs\cite{7,8} and the machine learning algorithms without the effective preprocessing\cite{14}. Our model has been applied with a very simple notion detect the face from the picture, cut it hereafter and then detect it with the trained model. The idea of yoloV3, yolov5 has been promoted recently but face mask recognition can be formulated with many simple notions than these.
\section{Conclusion}
In this paper, we have proposed an ultimate face cut algorithm that uses a pre-trained deep learning landmark detection model to detect facial 67 points. The 67 points are necessary to recognize all the facial landmarks that are useful for the face recognition task. Our technique uses only boundary points to take out those from the face and cut the face with the points so that we can avoid all the obstacles from the picture. The later stage has been carried out on the pre-processed dataset by the ResNet50 architecture. Although the ResNet50 architecture is very dense and heavy, we need to have some additional hardware requirements if we want to check the face mask from a video as the model will process the recognition frame by frame. Our proposed method have achieved an ACSA accuracy of 99.4 percent and loss 0.0183 on 20,721 data points that give a quite satisfactory result compared to the other models. We have also visualized some of the sample data with the Grad-CAM technique to understand the model feature points.

\section{Future Work}
The future work in this topic can be on the restoration of the face using PCA and other machine learning algorithms which can restore the face hidden behind the face mask. The engagement of Autoencoder can be a significant advancement but the only problem is the loss function of the autoencoder. The autoencoder gives a blurry image if the loss function is not working well. We can use MSE-SSIM loss for that issue. This face mask recognition technique can be highly dependent on just a subnetwork of ResNet50, as it is a very heavy architecture to determine the face mask task.

\vspace{12pt}
\color{red}

\begin{thebibliography}{00}
\bibitem{1} 
Hongzhou Lu, Charles W. Stratton, and Yi-Wei Tang. Outbreak of
pneumonia of unknown etiology in Wuhan, China: The mystery and
the miracle. Journal of Medical Virology, 92(4):401–402, 2020.

\bibitem{2} Chih-Cheng Lai, Tzu-Ping Shih, Wen-Chien Ko, HungJen Tang, and Po-
Ren Hsueh. Severe acute respiratory syndrome coronavirus 2 (SARS-
CoV-2) and coronavirus disease-2019 (COVID-19): The epidemic
and the challenges. International Journal of Antimicrobial Agents,
55(3):105924, March 2020.

\bibitem{3} Hussin A. Rothan and Siddappa N. Byrareddy. The epidemiology and
pathogenesis of coronavirus disease (COVID-19) outbreak. Journal of
Autoimmunity, page 102433, February 2020.

\bibitem{4} Sunny H. Wong, Jeremy Y. C. Teoh, Chi-Ho Leung, William K. K. Wu, Benjamin H. K. Yip, Martin C. S. Wong, and David S. C. Hui COVID-19 and Public Interest in Face Mask Use.
\bibitem{5} Md. Rafiuzzaman Bhuiyan, Sharun Akter Khushbu, Md. Sanzidul Islam A Deep Learning Based Assistive System to
Classify COVID-19 Face Mask for Human Safety
with YOLOv3 11th ICCNT 2020.
\bibitem{6} Tulin Ozturk, Muhammed Talo, Eylul Azra Yildirim, Ulas Baran
Baloglud, Ozal Yildirim, U. Rajendra Acharya , Automated detection
of COVID-19 cases using deep neural networks with X-ray images,
2020.
\bibitem{7} Joao Carlos Virgolino Soares, Marcelo Gattass, Marco Antonio Meggi-
olaro, “Visual SLAM in Human Populated Environments: Exploring the
Trade-off between Accuracy and Speed of YOLO and Mask R-CNN”,
19th International Conference on Advanced Robotics (ICAR), 2019.

\bibitem{8} Ceren Gulra Melek; Elena Battini Sonmez; Songul Albayrak, Object Detection in Shelf Images with YOLO. IEEE EUROCON 2019 -18th International Conference on Smart Technologies.

\bibitem{9} Arjya Das; Mohammad Wasif Ansari; Rohini Basak, Covid-19 Face Mask Detection Using TensorFlow, Keras and OpenCV,2020 IEEE 17th India Council International Conference (INDICON).

\bibitem{10} \url{https://github.com/prajnasb/observations/tree/master/experiements/data}.

\bibitem{11} \url{https://www.kaggle.com/omkargurav/face-mask-dataset}

\bibitem{12}
\url{https://www.kaggle.com/ashishjangra27/face-mask-12k-images-dataset}

\bibitem{13}
Authors:Garrett Wilson,Diane J Cook,
A Survey of Unsupervised Deep Domain Adaptation.\url{https://doi.org/10.1145/3400066}

\bibitem{14}
Wuttichai Vijitkunsawat, Peerasak Chantngarm, Study of the Performance of Machine Learning Algorithms for Face Mask Detection.

\bibitem{15}
Guanhao Yang, Wei Feng, Jintao Jin, Qujiang Lei, Xiuhao Li, Guangchao Gui, Weijun Wang, Face Mask Recognition System with YOLOV5 Based on Image Recognition.

\end{thebibliography}
\end{document}